\documentclass{article}

\usepackage{algorithm}
\usepackage{algpseudocode}
\usepackage{arxiv}
\usepackage{amsmath}
\usepackage[utf8]{inputenc} % allow utf-8 input
\usepackage[T1]{fontenc}    % use 8-bit T1 fonts
\usepackage{hyperref}       % hyperlinks
\usepackage{url}            % simple URL typesetting
\usepackage{booktabs}       % professional-quality tables
\usepackage{amsfonts}       % blackboard math symbols
\usepackage{nicefrac}       % compact symbols for 1/2, etc.
\usepackage{microtype}      % microtypography
\usepackage{lipsum}		% Can be removed after putting your text content
\usepackage{graphicx}
\usepackage{natbib}
\usepackage{doi}
\usepackage{amssymb}
\usepackage{multirow}
\usepackage{tabularx}
\usepackage[colorinlistoftodos]{todonotes}
\usepackage{bbm}
\usepackage{bm}

\title{Evaluating the Quality of the Quantified Uncertainty for (Re)Calibration of Data-Driven Regression Models}

\date{\today}	% Here you can change the date presented in the paper title
\author{ \href{https://orcid.org/0000-0002-0728-0071}{\includegraphics[scale=0.06]{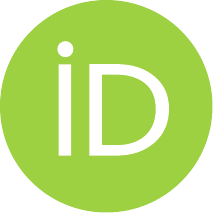}\hspace{1mm} Jelke Wibbeke}\\
	Department of Computing Science\\
		Carl von Ossietzky Universit{\"a}t Oldenburg\\
	Ammerl{\"a}nder Heerstraße 114-118,\\ 26129 Oldenburg, Germany\\
	\texttt{jelke.wibbeke@uni-oldenburg.de}\\
	\And
    \href{https://orcid.org/0009-0001-4487-1639}
    {\includegraphics[scale=0.06]{orcid.pdf}\hspace{1mm} Nico Sch{\"o}nfisch}\thanks{The work was carried out while affiliated with the Carl von Ossietzky Universit{\"a}t Oldenburg, Oldenburg, Germany}\\
    Institute of Systems Engineering for Future Mobility\\
    German Aerospace Center (DLR)\\
    Escherweg 2, 26121 Oldenburg, Germany\\
    \texttt{nico.schoenfisch@dlr.de}\\    
    \And
    \href{https://orcid.org/0009-0004-6875-7912}    
    {\includegraphics[scale=0.06]{orcid.pdf}\hspace{1mm}Sebastian Rohjans}\\
	Department for Civil Engineering, \\ Geoinformation and Health Technology\\
	Jade University of Applied Science\\
    Ofener Str. 16/19, 26121 Oldenburg, Germany\\
	\texttt{sebastian.rohjans@jade-hs.de}\\
	\And
	\href{https://orcid.org/0000-0002-1548-6547}{\includegraphics[scale=0.06]{orcid.pdf}\hspace{1mm}Andreas Rauh}\\
	Department of Computing Science\\
	Carl von Ossietzky Universit{\"a}t Oldenburg\\
	Ammerl{\"a}nder Heerstraße 114-118,\\ 26129 Oldenburg, Germany\\
	\texttt{andreas.rauh@uni-oldenburg.de}\\
}

\renewcommand{\shorttitle}{\textit{arXiv} Template}

\hypersetup{
pdftitle={Evaluating the Quality of the Quantified Uncertainty for (Re)Calibration of Data-Driven Regression Models},
pdfsubject={cd.LG},
pdfauthor={Jelke Wibbeke, Nico Sch{\"o}nfisch, Sebastian Rohjans, Andreas Rauh},
pdfkeywords={Uncertainty Quantification, Supervised learning, Ensembles, Likelihood, Review, Benchmark},
}

\begin{document}
\maketitle
This manuscript is a preprint version of the article that has since been peer-reviewed and published in the International Journal of Approximate Reasoning. The published version includes revisions and improvements resulting from the review process and should be considered the authoritative version of record. The final published article is available open access at: \url{https://doi.org/10.1016/j.ijar.2026.109685}.

\begin{abstract}
In safety-critical applications data-driven models must not only be accurate but also provide reliable uncertainty estimates. This property, commonly referred to as calibration, is essential for risk-aware decision-making. In regression a wide variety of calibration metrics and recalibration methods have emerged. However, these metrics differ significantly in their definitions, assumptions and scales, making it difficult to interpret and compare results across studies. Moreover, most recalibration methods have been evaluated using only a small subset of metrics, leaving it unclear whether improvements generalize across different notions of calibration. In this work, we systematically extract and categorize regression calibration metrics from the literature and benchmark these metrics independently of specific modelling methods or recalibration approaches. Through controlled experiments with real-world, synthetic and artificially miscalibrated data, we demonstrate that calibration metrics frequently produce conflicting results. Our analysis reveals substantial inconsistencies: many metrics disagree in their evaluation of the same recalibration result, and some even indicate contradictory conclusions. This inconsistency is particularly concerning as it potentially allows cherry-picking of metrics to create misleading impressions of success. We identify the Expected Normalized Calibration Error (ENCE) and the Coverage Width-based Criterion (CWC) as the most dependable metrics in our tests. Our findings highlight the critical role of metric selection in calibration research.

\end{abstract}

% keywords can be removed

\section{Introduction} \label{sec:intro}
Predictive models in machine learning are increasingly used in decision-making processes, particularly in high-stakes domains such as autonomous driving, medicine, and finance \cite{wang_uncertainty_2025, laves_well-calibrated_2020, kumbure_machine_2022}. In these applications, it is not sufficient for models to produce only accurate point predictions; they must also provide reliable estimates of their uncertainty. For instance, a self-driving car interpreting its environment must not only identify potential hazards but also express how confident it is in those interpretations. Such confidence estimates, or quantified uncertainties, are vital for making robust, risk-aware decisions.

In regression tasks, uncertainty is typically expressed as a predictive distribution over the target variable or as a prediction interval. Various methods have been proposed to generate uncertainty estimates, including Monte Carlo dropout \citep{gal16_dropout}, deep ensembles \citep{Lakshminarayanan_simple}, or Bayesian neural networks \citep{jospin_hands-on_2022}. These advances have spurred significant interest in uncertainty quantification (UQ) \citep{abdar_review_2021, kabir_neural_2018}, yet adoption in safety-critical real-world applications remains limited \citep{gawlikowski_survey_2023}. This limitation is partly due to the difficulty of evaluating the quality of the estimated uncertainty, especially when ground truth uncertainty is not available \citep{hullermeier_aleatoric_2021, sluijterman_howtoevaluate, gawlikowski_survey_2023}.

Crucially, we must distinguish between different yet related concepts: uncertainty quantification is the process of estimating uncertainty in predictions; quantified uncertainty is the measurable result of this process (e.g., standard deviation, interval width); and the quality of quantified uncertainty (QQU) describes how well these estimates reflect the true uncertainty in the data. Evaluating this quality is challenging, particularly in regression, where labels are continuous and uncertainty can stem from both aleatoric sources (inherent noise) and epistemic sources (lack of knowledge) \citep{hullermeier_aleatoric_2021}.

Many authors summarize both the accuracy of predictions and the QQU under the term calibration. In this view, a well-calibrated model not only predicts accurately but also assigns uncertainty estimates that reliably reflect the true variability in the prediction. However, the term calibration is used inconsistently. Oftentimes calibration is simultaneously used to refer to the process of improving a model’s uncertainty estimates \citep{Lakshminarayanan_simple, hullermeier_aleatoric_2021, he_asurvey_2025, psaros_uncertainty_2023, guo_calibration_2017, song_distribution_2019, cui_calibrated_2020}. This inconsistency, one talking about model performance and one about performance improvement, can be confusing, especially when discussing the evaluation or improvement of uncertainty estimates. 

\begin{figure}[htbp]
    \centering
    \begin{tikzpicture}[
        def/.style={
            draw,
            rounded corners,
            text width=0.4\textwidth,
            align=left,
            inner sep=8pt
        },
        struc/.style={
            draw,
            rounded corners,
            text width=0.4\textwidth,
            align=left,
            inner sep=8pt
        },
        node distance=0.25cm and 0.2cm
    ]
    % Left column - one large box
    \node[struc] (structure) {
        \textbf{Article Structure:}\\
        \ref{sec:intro} \nameref{sec:intro} \\
        \ref{sec:related_work} \nameref{sec:related_work}\\  
        \ref{sec:methods} \nameref{sec:methods}\\  
        \quad\ref{sec:picp} \nameref{sec:picp}\\  
        \quad\ref{sec:cwc} \nameref{sec:cwc}\\
        \quad\ref{sec:interval_score} \nameref{sec:interval_score}\\
        \quad\ref{sec:crps} \nameref{sec:crps}\\
        \quad\ref{sec:nll} \nameref{sec:nll}\\
        \quad\ref{sec:ece} \nameref{sec:ece}\\
        \quad\quad\ref{sec:cal_score} \nameref{sec:cal_score}\\  
        \quad\quad\ref{sec:ence} \nameref{sec:ence}\\
        \quad\quad\ref{sec:ecpe} \nameref{sec:ecpe}\\  
        \quad\quad\ref{sec:uce} \nameref{sec:uce}\\
        \quad\quad\ref{sec:qce} \nameref{sec:qce}\\
        \quad\ref{sec:comparability} \nameref{sec:comparability}\\  
        \ref{sec:benchmark} \nameref{sec:benchmark}\\  
        \quad\ref{sec:synthetic} \nameref{sec:synthetic}\\  
        \quad\ref{sec:real_world} \nameref{sec:real_world}\\ 
        \quad\ref{sec:calibration} \nameref{sec:calibration}\\
        \quad\ref{sec:miscalibration} \nameref{sec:miscalibration}\\
        \quad\ref{sec:recommendations} \nameref{sec:recommendations}\\
        \ref{sec:discussion} \nameref{sec:discussion}\\    
        \ref{sec:outlook} \nameref{sec:outlook}\\  
        \ref{sec:conclusion} \nameref{sec:conclusion}
    };
    % Right column - two stacked boxes
    \node[def, right = of structure.north east, anchor=north west] (one) {
        \textbf{Accuracy/Error:}\\
        \emph{Accuracy} or \emph{Error} refers to how close the predicted values are to the true target values. It is typically measured using metrics such as mean squared error (MSE) or mean absolute error (MAE).
    };
    \node[def, below = of one] (three) {
        \textbf{Uncertainty:}\\
        \emph{Uncertainty} refers to the model’s estimate of how much its prediction might deviate from the true value. It reflects the confidence in a prediction and can be split in \emph{aleatoric} and \emph{epistemic uncertainty}.
    };
    \node[def, below= of three] (four) {
        \textbf{Aleatoric Uncertainty}\\
        \emph{Aleatoric uncertainty} originates from inherent variability in the data, such as measurement noise. It cannot be reduced by collecting more data and is typically modeled as input-dependent noise in the output distribution.
    };
    \node[def, below= of four] (five) {
        \textbf{Epistemic Uncertainty}\\
        \emph{Epistemic uncertainty} originates from a lack of knowledge about the modeled system or insufficient model structure. It reflects ignorance and can be reduced by collecting more representative data or using more expressive models.
    };
    \node[def, below= of structure] (six) {
        \textbf{Uncertainty Quantification (UQ)}\\
        \emph{Uncertainty quantification} is the process of estimating uncertainty in model predictions. 
    };
    \node[def, below= of six] (seven) {
        \textbf{Quantified Uncertainty}\\
        \emph{Quantified uncertainty} is the measurable output of an \emph{uncertainty quantification} method, such as a standard deviation, interval, or distribution associated with a prediction.
    };
    \node[def, below= of seven] (eight) {
        \textbf{Quality of Quantified Uncertainty (QQU)}\\
        \emph{Quality of quantified uncertainty} refers to how well the \emph{quantified uncertainty} reflects the true variability and confidence in the model's predictions.
    };
    \node[def, below= of five] (nine) {
        \textbf{Calibration}\\
        \emph{Calibration} refers to the extent to which a model’s \emph{quantified uncertainty} and \emph{accuracy} align with the actual variability in the outcomes.
    };
    \node[def, below= of nine] (ten) {
        \textbf{Recalibration}\\
        \emph{Recalibration} is the process of adjusting a model's predicted uncertainty in order to improve its \emph{calibration}.
    };
    \end{tikzpicture}
    \caption{Structure and taxonomy of the article. The full metric names are provided in Section~\ref{sec:methods}~\nameref{sec:methods}.} \label{fig:taxonomy}
\end{figure}
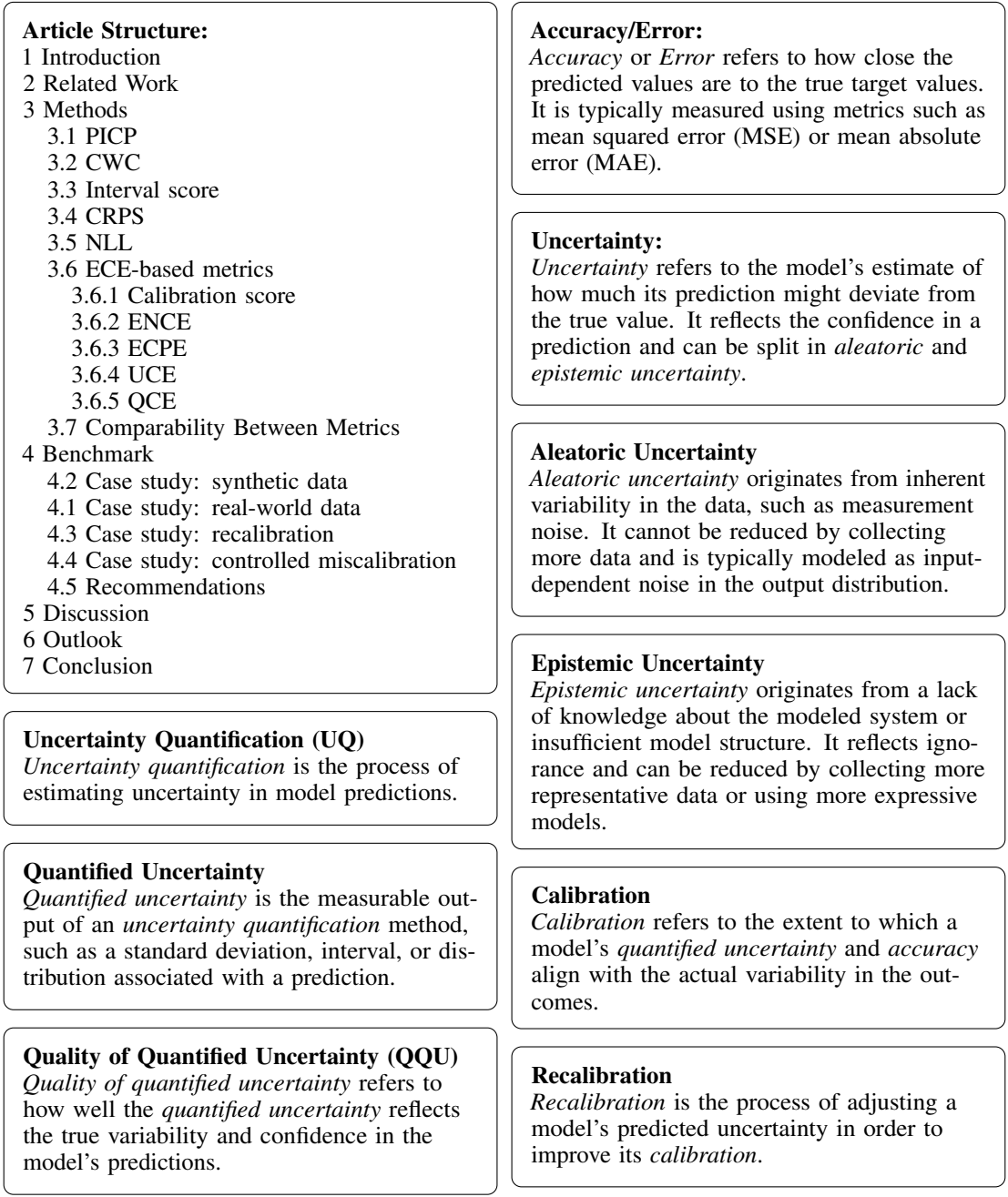

In this work, we adopt the more precise terminology which defines calibration as the state of having accurate predictions and correct uncertainty estimates, and refers to the process of improving said state as recalibration \citep{gawlikowski_survey_2023, levi_evaluating_2022, dheur_large-scale_2023, kuleshov_accurate_2018, laves_well-calibrated_2020, kuppers_parametric_2023}. Under this taxonomy, recalibration methods aim to improve model calibration, which is the state of providing accurate predictions and correct uncertainty estimates. To do so, recalibration methods relying on calibration metrics to assess and quantify the success. A comprehensive overview of the taxonomy used in this work is provided in Fig.~\ref{fig:taxonomy}.

An essential component of any recalibration method is a calibration metric to evaluate the accuracy and QQU. Without such a metric, the effect of recalibration cannot be assessed. Yet, here lies a critical issue: numerous metrics exist, but they differ widely in definition, scaling, and interpretation. For instance, the Quantile Calibration Error (QCE) by \citet{kuppers_parametric_2023} and the Calibration Score (CalS) by \citet{kuleshov_accurate_2018} both aim to assess model calibration (recalibration performance) but follow completely different approaches and yield values that are not directly comparable. This may not be problematic when evaluating a single model before and after recalibration, as only the metric difference is of interest. However, it becomes a considerable limitation when trying to compare different recalibration methods or benchmark models across datasets, since practitioners may be misled into preferring one method over another based solely on the choice of metric rather than true improvements in calibration.

Metrics to measure the calibration are embedded in all recalibration studies. However, while recalibration methods are routinely benchmarked against one another, the metrics they rely on are rarely, if ever, compared. This leaves open a critical question: how much of a recalibration method’s perceived performance is actually due to the underlying calibration metric used in its evaluation? Without a direct comparisons of the metrics themselves, the influence of the chosen metric remains opaque.

Thus, despite a growing body of research on uncertainty estimation and recalibration, a fundamental gap remains: there has been no systematic comparison of metrics used to evaluate calibration. This lack of comparability hinders progress in the field and makes it difficult to assess which recalibration methods or UQ approaches are truly effective. Several recent reviews highlight this challenge and identify improved recalibration and evaluation techniques as key research gaps \citep{abdar_review_2021, psaros_uncertainty_2023, wang_uncertainty_2025}.

To move beyond this gap, we take a different perspective: we shift focus to the evaluation itself. All recalibration approaches rely (explicitly or implicitly) on a calibration metric to measure the performance improvement before and after recalibration. Yet these metrics are rarely scrutinized themselves.

In this work, we systematically extract a broad range of calibration metrics and evaluate them independently of any specific model or recalibration method. Our benchmark is conducted in a model-agnostic setting, where we assume access to a set of predictions and corresponding quantified uncertainties for each data point. This isolates the behavior of the metrics themselves from the underlying modeling or uncertainty estimation approach (e.g., ensembling, Bayesian inference, or MC dropout).

A central reason why these metrics have not been compared before lies in their incompatible mathematical foundations and scaling: some are probabilistic, others interval-based; some return scores that are bounded, others unbounded; and their absolute values are not directly comparable. To address this, we introduce an evaluation framework that compares metrics by analyzing how they rank the calibration across different datasets. The intuition is simple: if a model is well calibrated, its performance should be ranked similarly across all metrics. If different metrics yield conflicting rankings -- for instance, one judging a dataset’s calibration as high while another deems it low -- this reveals a fundamental inconsistency in the metrics.

This article makes the following key contributions:

\begin{itemize}
    \item We extract and categorize a comprehensive set of metrics from the literature that are used to evaluate the calibration in regression.
    \item We benchmark these metrics in a controlled experimental setting using both synthetic and real-world datasets.
    \item We show that calibration metrics often disagree in their assessments, which underscores the critical impact of metric choice when evaluating or comparing recalibration methods.
    \item We show that the Expected Normalized Calibration Error (ENCE) is best suitable for assessing the calibration of a model.
\end{itemize}

An overview of the article structure including all metrics considered in the benchmark is provided in Fig.~\ref{fig:taxonomy}. The sections are arranged as follows: an overview over the related work is given in Section~\ref{sec:related_work}. In Section~\ref{sec:methods} we present the extracted metrics. In Section~\ref{sec:benchmark} the quantitative benchmark using synthetic and real-world data is shown, concluding with application recommendations. This is followed by a discussion of potential limitations of our benchmark in Section~\ref{sec:discussion}. To conclude the article we present an outlook on possible implications and future work in Section~\ref{sec:outlook} and a conclusion with summary in Section~\ref{sec:conclusion}.

\section{Related Work}\label{sec:related_work}
In classification tasks, the evaluation of calibration quality is well-established, with the Expected Calibration Error (ECE) being one of the most widely used metrics \citep{guo_calibration_2017, gawlikowski_survey_2023}. The ECE estimates how well predicted probabilities reflect empirical outcomes by binning confidence scores and comparing them to empirical accuracy. However, transferring this concept to regression is not straightforward. In regression, predicted uncertainties are continuous and typically take the form of prediction intervals or distributions.

Some authors have proposed workarounds to apply classification-style calibration metrics to regression. For instance, \citet{chung_beyond_2021} uses pinball loss to discretize the target space, effectively reducing the regression task to a pseudo-multiclass setting where the ECE becomes applicable. However, such transformations can distort the nature of the regression uncertainty and raise questions about metric interpretability.

In addition, adaptations of ECE have been proposed, such as the the Calibration Score (CalS) by \citet{kuleshov_accurate_2018}. \citet{kuleshov_accurate_2018} introduce one of the first recalibration approaches for regression models called Isotonic Regression, proposing the CalS as a metric to measure success in combining prediction accuracy and QQU. Both are widely used in (re)calibration applications. However, \citet{levi_evaluating_2022} highlight limitations of the CalS and instead introduce recalibration by variance scaling and the Expected Normalized Calibration Error (ENCE) as a metric. While both metrics are commonly included in recalibration software, their relative strengths and weaknesses remain largely unexplored.

Other metrics have emerged in specific modeling contexts. The Continuous Ranked Probability Score (CRPS) is used to evaluate probabilistic forecasts in large-scale regression settings \citep{dheur_large-scale_2023, morris_neural_2023}. The CRPS, which integrates both the QQU and accuracy, is often used alongside the Negative Log-Likelihood (NLL), though authors refrain from ranking one over the other. These metrics are typically employed as training objectives rather than calibration evaluation metrics.

Point-based metrics such as the Prediction Interval Coverage Probability (PICP) and the Prediction Interval Width (PIW) are also commonly used to assess the quality of prediction intervals (PIs). However, these metrics can be misleading when used in isolation. For example, maximizing coverage of the PICP can be trivially achieved by widening intervals. To mitigate this, \citet{khosravi_comprehensive_2011} propose the Coverage Width-based Criterion (CWC), which incorporates both coverage and width of the interval. 

Simulation-based approaches provide alternative means of evaluating calibration. \citet{sluijterman_howtoevaluate} propose a method to estimate point-wise coverage probability, which requires knowledge of the data-generating function and is therefore limited to synthetic settings. Similarly, the Confidence Interval Coverage Probability (CICP) has been applied in simulated environments where the true function is known. These simulation-based approaches provide the advantage of knowing the ground-truth uncertainty. However, the need for data-generating functions severely limits applicability to real-world settings.

While many methods are proposed to construct uncertainty estimates -- ranging from Bayesian neural networks \citep{jospin_hands-on_2022}, Deep Ensembles \citep{Lakshminarayanan_simple}, and MC Dropout \citep{gal16_dropout} -- these works generally focus on algorithmic techniques. However, the construction of uncertainty estimates is not the focus of this work and we refer to the respective review studies, for example by \citet{kabir_neural_2018}, \citet{gawlikowski_survey_2023} or \citet{he_asurvey_2025}. Benchmarks of construction methods often emphasize empirical success on specific datasets rather than offering systematic comparisons of the underlying evaluation metrics. Recent surveys \citep{gawlikowski_survey_2023, he_asurvey_2025} highlight this gap, noting that even though various uncertainty quantification methods and recalibration approaches are available, analysis of the metrics used to evaluate them is lacking.

In summary, while classification tasks benefit from a dominant and interpretable evaluation framework for calibration, the regression domain lacks consensus. Existing calibration metrics differ in assumptions, scales, and target formulations, making it difficult to compare methods across studies. This motivates the need for a systematic, model-agnostic comparison of calibration metrics for regression.

A first step in this direction is taken by \citet{kristoffersson_lind_uncertainty_2024} indicating that the Calibration Score metric by \citet{kuleshov_accurate_2018} is most stable and interpretable. However, the scope of the review is limited as only four metrics are compared using four synthetic datasets. In our review we extend the scope and include 13 metrics, evaluating the performance on in total 26 datasets.

\section{Methods} \label{sec:methods}
In this section, we present various model-agnostic metric to assess the calibration of regression models. In addition, we also include some metrics that purely measure the QQU. We aim at model-agnostic metrics in that regard that the metrics can be applied independently of the method that was used to construct the uncertainty estimate. A brief summary of all presented metrics is provided in Tab. \ref{tab:metrics} of Section \ref{sec:comparability}.

Without claiming to be exhaustive, we cover the most popular metrics in current use, which are:
\begin{itemize}
    \item Prediction Interval Coverage Probability (PICP)
    \item Coverage Width-based Criterion (CWC)
    \item Interval Score (IS)
    \item Continuous Ranked Probability Score (CRPS)
    \item Negative Log-Likelihood (NLL)
    \item Calibration Score (CalS)
    \item Normalized Calibration Error (ENCE)
    \item Expectation of Coverage Probability Error (ECPE)
    \item Uncertainty Calibration Error (UCE)
    \item Quantile Calibration Error (QCE)
\end{itemize}

Regarding the mathematical notation, we consider a dataset consisting of $N$ samples with corresponding target values and features. We denote the dataset as $\mathbf{D}=(\mathbf{y}, \mathbf{X})$, where $\mathbf{y} \in \mathbb{R}^N$ is the target vector $\mathbf{y} = [y_1, y_2, \cdots, y_N]^\intercal$, containing the target value $y_i$ for each sample $i = 1, \dots, N$. The feature matrix $\mathbf{X} = [\mathbf{x}_1, \mathbf{x}_2, \cdots, \mathbf{x}_N]^\intercal \in \mathbb{R}^{N \times d}$ contains the $d$-dimensional feature vectors $\mathbf{x}_i$ for each sample. Each individual sample is thus represented by a pair $(y_i, \mathbf{x}_i)$, where $y_i \in \mathbb{R}$ and $\mathbf{x}_i \in \mathbb{R}^d$. 

The model $M$ outputs per sample $\mathbf{x}_i$ either a prediction interval $\text{PI}_i = [\hat{y}_{\text{l},i}, \hat{y}_{\text{u},i}]$ enclosing the true value or the parameters of an estimated probability density function (PDF) $\hat{f}_i$, which in case of an assumed Gaussian distribution are the mean $\hat{\mu}_i = \hat{y}_i$ approximating the true target and the standard deviation $\hat{\sigma}_i$. In addition, we denote $\hat{F}_i$ to be the cumulative density function of the estimated distribution $\hat{f}_i$.

This notation will be used throughout the following sections.

\subsection{PICP} \label{sec:picp}
The Prediction Interval Coverage Probability (PICP) is one of the most common and simplest possibilities to evaluate the quality of the uncertainty estimates of a regression model. It describes the fraction of observations that are enclosed inside the corresponding PIs according to 

\begin{equation}
    \text{PICP} = \frac{1}{N} \sum_{i=1}^{N} \mathbbm{1}\{\hat{y}_i \in \text{PI}_i\}.
\end{equation}
The predictions interval $\text{PI}_i = [\hat{y}_{\text{l},i}, \hat{y}_{\text{u},i}]$ contains a lower and an upper bound of the predicted value.  Interpreting the PICP is dependent on the chosen prediction intervals, as an arbitrarily high PICP can be archived by widening the PI. Ideally, the PICP should be very close or marginally larger than the nominal confidence level associated to the PIs. A common choice is a 95\% prediction interval. In that case, the lower and upper bounds of the uncertainty interval are chosen in a way that ideally 95\% of the target values are enclosed. If the PICP is lower than the associated confidence level the model is over-confident, whereas much higher values indicate under-confidence.\footnote{Explanation as this might come across as counterintuitive: If fewer predictions fall into the PI as intended that means that the PI is too narrow. The model is too sure of its predictions and thus over-confident.} However, a good PICP does not guarantee point-wise coverage and a good marginal coverage does not imply a good point-wise coverage~\citep{sluijterman_howtoevaluate}. Being the fraction of observations, the PICP can take on values between 0 and 1.

Albeit not being used as a metric to assess model calibration, the PICP is commonly used to make a statement about a model's QQU, as it is easy to compute and interpret.

\subsection{CWC}\label{sec:cwc}
The Coverage Width-based Criterion (CWC) introduced by \citet{khosravi_comprehensive_2011} aims to eliminate the shortcoming of the PICP in that it only considers whether the prediction lies within the interval, but not how wide the interval is. This means that an arbitrarily high PICP can be achieved if the interval is chosen wide enough. To take this into account, the CWC contains both the PICP and the normalized mean PI width (NMPIW).

The NMPIW is calculated according to
\begin{equation}
    \text{NMPIW} = \frac{\text{MPIW}}{R},
\end{equation}
where $R$ denotes the range of the target variable and the mean PI width (MPIW) is quantified according to
\begin{equation}
    \text{MPIW} = \frac{1}{N} \sum^{N}_{i=1} (\hat{y}_{\text{u},i} - \hat{y}_{\text{l},i}).
\end{equation}

Ultimately, \citet{khosravi_comprehensive_2011} define the CWC according to
\begin{equation}
    \text{CWC} = \text{NMPIW} \cdot \left(1 + \mathbbm{1}\{\text{PICP} < \lambda\}\cdot e^{-\eta\cdot\left(\text{PICP} - \lambda\right)}\right),
\end{equation}
where $\eta$ is a hyperparameter, $\mathbbm{1}\{\cdot\}$ the indicator function and $\lambda$ corresponds to the nominal confidence level associated with the PI.

The indicator function $\mathbbm{1}\{\text{PICP} < \lambda\}$ returns zero and thus eliminates the exponential term of the CWC if the PICP is greater or equal to the nominal confidence level $\lambda$, leaving only the NMPIW. Thus, the CWC can be minimized by minimizing the NMPIW as long as the PICP is greater than the nominal confidence level $\lambda$ resulting in tighter PIs. However, if the PICP$<\lambda$ the exponential term is larger than the NMPIW shifting the focus of the metric to the coverage probability. The more the PICP moves away from $\lambda$ the more the CWC increases, penalizing for violation of the coverage probability. The hyperparameter $\eta$ scales this penalization.
Theoretically, the values of the CWC can range from 0 to infinity, with lower values indicating better performance. The minimum value of zero is achieved when the NMPIW equals 0, which corresponds to a Dirac impulse. The maximum values can scale towards infinity depending on the choice of $\eta$, which determines the severity of the penalty for not meeting the PICP requirement. In their article, \citet{khosravi_comprehensive_2011} propose using $\eta = 50$.

\subsection{Interval score} \label{sec:interval_score}
The Interval Score (IS) is a proper scoring rule for evaluating PIs \citep{gneiting_strictly_2007}. It assesses the quality of uncertainty estimates by jointly considering both the width and whether the true value lies within it. In contrast to simple coverage metrics like the PICP, the IS penalizes both excessive width and non-coverage, thus encouraging sharp and well-calibrated intervals.

For a central $(1 - \alpha)$ PI of sample $i$ with confidence level $\alpha$ the IS is defined according to
\begin{equation}
\text{IS}_\alpha(\hat{y}_{\text{l},i}, \hat{y}_{\text{u},i}, y_i) = (\hat{y}_{\text{u},i} - \hat{y}_{\text{l},i}) + \frac{2}{\alpha} (\hat{y}_{\text{l},i} - y_i) \cdot \mathbbm{1}\{y_i < \hat{y}_{\text{l},i}\} + \frac{2}{\alpha} (y_i - \hat{y}_{\text{u},i}) \cdot \mathbbm{1}\{y_i > \hat{y}_{\text{u},i}\},
\end{equation}
where $\mathbbm{1}\{\cdot\}$ denotes the indicator function. The first term corresponds to the width of the interval, which encourages sharpness. The second and third term introduce penalties when the observation lies below or above the interval, respectively, with the penalty increasing proportionally to the distance from the interval bounds.

The interval score thus balances two competing goals: minimizing the interval width and ensuring that the interval contains the true value. Predictions that are too narrow and miss the target are penalized more severely than those that are wide but correctly include the target.

The average interval score over all $N$ samples is given by
\begin{equation}
\text{IS}_{\alpha,\text{avg}} = \frac{1}{N} \sum_{i=1}^{N} \text{IS}_\alpha(\hat{y}_{\text{l},i}, \hat{y}_{\text{u},i}, y_i).
\end{equation}

The interval score is strictly proper, meaning that it is minimized in expectation when the predicted interval matches the true predictive distribution. It has the same unit as the target variable and takes values in the range $[0, \infty)$, where lower scores indicate better predictive performance.

\subsection{CRPS} \label{sec:crps}
The Continuous Ranked Probability Score (CRPS) is a generalization of the Brier Score to regression, which measures the accuracy of probabilistic predictions in a classification context \citep{gneiting_strictly_2007}. As the Brier Score, the CRPS is a proper scoring rule in that the minimum expected score is archived if the true distribution is predicted. The CRPS is defined according to
\begin{equation}
\text{CRPS}(\hat{F}_i, y_i) = \int_{-\infty}^{\infty} (\hat{F}_i(a) - \mathbbm{1}\{z \geq y_i\})^2 \mathrm{d}z
\end{equation}
where $\hat{F}_i$ represents the cumulative distribution function (CDF) of the predicted probability distribution $\hat{f}_i$ of sample $i$. The variable $z$ is a variable of integration over the range of $y$. 

The CRPS is the generalization the absolute error of the prediction to which it reduces if the prediction is not a distribution $\hat{f}_i$ but a point measure $\hat{y}_i$ (not probabilistic) \citep{gneiting_strictly_2007}. It is calculated as the integral over the target domain between the predicted CDF and the indicator function $\mathbbm{1}\{y \le z\}$, which represents the “true” CDF for each prediction. The CRPS measures the squared difference between these two curves across all possible target values, on a per-sample basis. It has the same unit as the target variable and can range from $0$ to $+\infty$, where smaller values indicate better performance. In short: the CRPS tells how close the PDF is to reality by comparing it to the ideal step-shaped distribution of the truth. The smaller the CRPS, the sharper and better aligned the predictions are with the actual outcome.

With the assumption that the predicted distribution is Gaussian with mean $\hat{\mu}$ and variance $\hat{\sigma}^2$ the integral can be expressed in closed form as
\begin{equation}
\text{CRPS}(\mathcal{N}(\hat{\mu}_i, \hat{\sigma}_i^2), y_i) = \hat{\sigma}_i \left[ \frac{1}{\sqrt{\pi}} - 2 \phi \left( \frac{y_i - \hat{\mu}_i}{\hat{\sigma}_i} \right) - \frac{y_i - \hat{\mu}_i}{\hat{\sigma}_i} \left( 2 \Phi \left( \frac{y_i - \hat{\mu}_i}{\hat{\sigma}_i} \right) - 1 \right) \right],
\end{equation}
where $\phi$ denotes the probability density and $\Phi$ the CDF of a standard Gaussian variable.

The CRPS of multiple samples is calculated as the mean of the individual scores.

\subsection{NLL} \label{sec:nll}
The Negative Log-Likelihood (NLL) measures the likelihood of the observed outcomes over the model's predicted probability distribution and therefore how well the predicted probability distribution explains the observed data. Thus, minimizing the NLL is equal to maximizing the likelihood of the observed data \citep{gawlikowski_survey_2023}. The NLL can be obtained according to
\begin{equation}
    \text{NLL}\left(\mathbf{\hat{y}}, \mathbf{x}\right) = - \sum^N_{i=1} \mathrm{log}\left(\hat{f}\left(\hat{y}_i|\mathbf{x}_i\right)\right),
\end{equation}
where $\hat{f}\left(\hat{y}_i|\mathbf{x}_i\right)$ denotes the predicted PDF.

To use the NLL in a regression setting an assumption about the distribution of the prediction has to be made. Considering a Gaussian distribution (for simplicity and in absence of arguments in favor of a specific other distribution) the NLL can be used for regression as in \citet{morris_neural_2023} according to 
\begin{equation}
    \text{NLL}\left(\mathbf{y}, \mathbf{\hat{y}}, \boldsymbol{\hat{\sigma}}\right) = \frac{1}{N} \sum_{i=i}^N \left(\frac{1}{2 \hat{\sigma}^2_i}\left(y_i - \hat{y_i}\right)^2 + \frac{1}{2} \mathrm{log}\left(2\pi\hat{\sigma}_i^2\right)\right).
\end{equation}
Here, $\mathbf{y}$ contains the ground truth target values and $\mathbf{\hat{y}}$ and $\boldsymbol{\hat{\sigma}}$ are the model's predictions with the associated standard deviations.

The NLL penalizes incorrect predictions (via the squared-error term) as well as over- or under-confidence in predictions (via the log-variance term). It can take on values between negative and positive infinity, where a smaller value indicates a better model. However, assessing the NLL value in isolation provides little insight because its absolute value depends on factors like data scale and predicted uncertainty, making it meaningless without comparison to other NLL values using the same probabilistic setup. Therefore, the NLL is often used as a recalibration metric (before and after recalibration) or as a loss function during model training.

\subsection{ECE-based metrics} \label{sec:ece}
The Expected Calibration Error (ECE) is a prominent measure to assess the calibration (and thus the quality of the quantified uncertainty) in the classification setting \citep{gawlikowski_survey_2023,guo_calibration_2017}. The idea behind the ECE is to group the predicted confidence scores of all samples (typically the probability assigned to the predicted class) into bins covering equal ranges of the confidence interval $[0, 1]$. For each bin, the average predicted confidence is compared to the empirical accuracy in that bin, i.e., the fraction of samples whose predicted class matches the true class. The closer the predicted confidences are to the corresponding empirical accuracies, the better the calibration. The ECE for the entire dataset is then computed as the weighted average of these absolute bin-wise differences, with weights proportional to the number of samples in each bin.

However, this standard way of using the ECE is specifically designed for classification tasks, because it explicitly compares the predicted probabilities of different classes to the empiric outcomes. As the knowledge gained from the ECE is also desirable for regression tasks, different adaptions of the ECE for the regressions setting have been made \citep{levi_evaluating_2022, kuleshov_accurate_2018,laves_well-calibrated_2020, zelikman_crude_2021, cui_calibrated_2020,chung_beyond_2021}. None of the approaches used the metric specifically to measure the calibration, but to assess the success of a recalibration. The two most prominent approaches being the expected normalized calibration error (ENCE) by \citet{levi_evaluating_2022} and the Calibration Score (CalS) by \citet{kuleshov_accurate_2018}. In the following, various approaches are presented.

\subsubsection{Calibration score} \label{sec:cal_score}
To asses the calibration of a model \citet{kuleshov_accurate_2018} propose the Calibration Score (CalS), which is an adaptation of the ECE to the regression setting. It combines the assessment of the prediction accuracy with the QQU based on the notion that for a well calibrated model the empirical and predicted CDFs should match as the dataset size goes to infinity. The CalS measures whether the predicted uncertainty intervals (e.g., 90\% prediction intervals) actually contain the true target values about the correct percentage of times.

To assess the quality, $m$ confidence thresholds $p_j$ are chosen with $0\leq p_1\leq p_2\dots \leq p_m \leq 1$ and for each threshold the empirical frequency is calculated according to
\begin{equation}
    \hat{p}_j = \frac{|\{y_i | F_i(y_i) \leq p_j, i = 1, \dots, N \}|}{N},
\end{equation}
with $F_i$ being the CDF of the target variable $y$. The CalS is then assessed according to 
\begin{equation}
    \text{CalS}(F,y) = \sum_{j=1}^m w_j \cdot (p_j - \hat{p}_j)^2,
\end{equation}
where $w_j$ is a weighting factor which can be chosen in dependence of the sample frequency to decrease the importance of intervals that contain fewer points. Thus, the calibration is evaluated by comparing the true frequency of points in each confidence interval relative to the predicted fraction of points in that interval. The CalS can take on values between 0 and the number of confidence thresholds $m$ (assuming all $w_j = 1$), where a lower score indicates a better performance.

To normalize the score, \citet{zelikman_crude_2021} proposed to use the RMSE instead of the cumulative squared error according to
\begin{equation}
    \text{CalS}_{\text{RMSE}}(F,y) = \sqrt{\frac{1}{m}\sum_{j=1}^m (p_j - \hat{p}_j)^2}.
\end{equation}
This way the CalS$_\text{RMSE}$ is bound between 0 and 1.

\citet{levi_evaluating_2022} argue that the CalS can be misleading, as it only evaluates a limited portion of the predictive distribution and can fail to reflect proper quality in the tails. Moreover, they demonstrate that models with visibly poor performance in reliability diagrams can still achieve low CalS values. To mitigate these issues \citet{levi_evaluating_2022} propose the ENCE as an improved version of the CalS.

\subsubsection{ENCE} \label{sec:ence}
The Expected Normalized Calibration Error (ENCE) is an adaption of the ECE for regression proposed by \citet{levi_evaluating_2022}. It follows the idea that a model is properly calibrated, if for each value of uncertainty the measured mean squared error of the prediction $\hat{y}_i$ matches the predicted variance $\sigma_i^2$. In short hand, this means that the root mean variance and root mean squared error are assumed to be equal.

For the regression setting, similar to the regular ECE, $N_\text{bin}$ bins are created. However, the binning is done based on the standard deviations of the model predictions for each sample. This means that predictions with similar standard deviations are sorted into the same bin. The ENCE is calculated over all $N_\text{bin}$ bins according to
\begin{equation}
    \text{ENCE} = \frac{1}{N_{\text{bin}}} \sum_{j=1}^{N_\text{bin}} \frac{|\text{RMV}(j) - \text{RMSE}(j)|}{\text{RMV}(j)},
\end{equation}
where $\text{RMV}(j)$ and $\text{RMSE}(j)$ refer to the root mean variance and root mean squared error of the respective bin $j$ according to
\begin{equation}
    \text{RMV}(j) = \sqrt{\frac{1}{|B_j|}\sum_{i \in B_j} \sigma_i^2}
\end{equation}
and
\begin{equation}
    \text{RMSE}(j) = \sqrt{\frac{1}{|B_j|}\sum_{i \in B_j} (y_i - \hat{y}_i)^2}.
\end{equation}
For both, $|B_j|$ is the number of elements in the $B_j$-th bin and $i$ the sample index.

The total number of bins can be chosen individually, ideally dividing the total number of samples to create equally sized bins \citep{levi_evaluating_2022}. Due to the division by the root mean variance the ENCE is normalized per bin, allowing to calculate the mean over the bins. The model achieves a good calibration quality if the RMSE is approximately equal to the RMV in each bin. By including variance and prediction error the ENCE captures sharpness of the QU as well as accuracy. The ENCE is strictly positive, where a smaller value indicates a better prediction quality.

\subsubsection{ECPE} \label{sec:ecpe}
The Expectation of Coverage Probability Error (ECPE) proposed by \citet{cui_calibrated_2020} combines the idea of using discrete confidence thresholds as in CalS with the PICP.
It measures the average absolute difference between the nominal confidence levels (i.e., the expected coverage probabilities for PI) and the empirical coverage observed in the data.

Formally, given a set of $m$ nominal confidence levels $p_j$ (e.g., 50\%, 80\%, 90\%, etc.) the respective PIs are constructed for each confidence level. The PICP is then computed as the fraction of the true target values that fall within said PIs. The ECPE measures the average absolute difference between the nominal confidence levels $p$ and the PICP according to

\begin{equation}
\mathrm{ECPE} = \frac{1}{m} \sum_{j=1}^m \left| p_j - \text{PICP}_j \right|.
\end{equation}
Thus, the EPCE is a more general form of the PICP leveraging the discretizing idea of the ECE-based metrics to evaluate the PICP at multiple confidence levels.

A lower ECPE indicates better QQU, meaning that the observed frequency of true values inside the prediction intervals closely matches the expected confidence levels. It can take on values between 0 and 1.

\subsubsection{UCE}\label{sec:uce}
The Uncertainty Calibration Error (UCE) proposed by \citet{laves_well-calibrated_2020} assumes that a good calibration is achieved if the values of uncertainty increase together with the prediction error. This is closely related to the ENCE by \citet{levi_evaluating_2022}.

To calculate the UCE, the range of the quantified uncertainty is partitioned into $N_\text{bin}$ bins with equal width. The average difference between the predictive error and the uncertainty is calculated weighted by the number of samples per bin according to 
\begin{equation}
    \text{UCE} = \sum_{j=1}^{N_\text{bin}}  \frac{|B_j|}{N}|\text{err}(B_j) - \text{uncert}(B_j)|,
\end{equation}
where $B_j$ is the set of samples indices for which the uncertainty falls into the $j$-th bin and $N = \sum_{j=1}^{N_\text{bin}} |B_j|$ is the total number of samples. The error per bin is calculated similar to the mean squared error 
\begin{equation}
    \text{err}(B_j) = \frac{1}{|B_j|} \sum_{i \in B_j} (y_i - \hat{y_i})^2.
\end{equation}
The uncertainty per bin is calculated using the mean variance according to
\begin{equation}
    \text{uncert}(B_j) = \frac{1}{|B_j|} \sum_{i \in B_j} \hat{\Sigma}_i^2,
\end{equation}
where $\hat{\Sigma^2}$ is the estimated uncertainty value, which in the case of an assumed Gaussian distribution is equal to the variance~$\hat{\sigma}^2$.

The UCE can take on values between 0 and infinity where a better calibration is indicated by lower UCE values.

The important distinction between the UCE and the ENCE is that the UCE uses the MSE instead of the RMSE and the mean variance instead of the root mean variance. This way the UCE is of a different scale than the target variable. In addition, the UCE it not normalized by the root mean variance per bin, which could potentially skew the UCE towards bins with larger uncertainty values. All in all, this makes the UCE more susceptible to outliers.

\subsubsection{QCE} \label{sec:qce}
The Quantile Calibration Error (QCE) by \citet{kuppers_parametric_2023} is inspired by the normalized estimation error squared (NEES) used in state estimation and Kalman filters to assess consistency. It measures the deviation between the nominal quantile level $\tau$ and the empirical coverage observed in model predictions. 

For a univariate regression model $M(\mathbf{x}_i)$ that predicts a mean $\hat{y}_i$ and a standard deviation $\hat{\sigma}_i$ of sample $i$, the NEES is defined as the squared Mahalanobis distance between the predicted and actual values, normalized by the model’s predicted uncertainty according to
\begin{equation}
    \epsilon_i = \frac{(y_i - \hat{y}_i)^2}{\hat{\sigma_i}^2}.
\end{equation}
Assuming a Gaussian predictive distribution and correct uncertainty, the NEES follows a chi-squared distribution with one degree of freedom, $\chi^2_1$. The corresponding threshold for quantile level $\tau$ is given by
\begin{equation}
    \alpha_\tau = F^{-1}_{\chi^2_1}(\tau),
\end{equation}
where $F^{-1}_{\chi^2_1}$ denotes the inverse CDF of the $\chi^2_1$ distribution.

The data is grouped into $N_{\text{bin}}$ bins according to the predicted standard deviation $\hat{\sigma_i}$, each containing $|B_j|$ samples, where $B_j$ is the set of sample indices in the respective bin. The QCE compares the fraction of samples for which $\epsilon_i \leq \alpha_\tau$ to the expected quantile level $\tau$. The empirical frequency in bin $j$ is
\begin{equation}
    \text{freq}(j) = \frac{1}{|B_j|} \sum_{i \in B_j} \mathbbm{1} \left( \epsilon_i \leq \alpha_\tau \right)
\end{equation}
and the QCE is estimated according to
\begin{equation}
    \text{QCE}(\tau) \approx \sum_{j=1}^{N_{\text{bin}}} \frac{|B_j|}{N} \left| \text{freq}(j) - \tau \right|,
\end{equation}
with $N$ being the total number of samples.

The QCE can take on values between 0 and 1, where a model with high accuracy and QQU will show $\text{freq}(j) \approx \tau$ in all bins, resulting in a low QCE.

\subsection{Comparability Between Metrics} \label{sec:comparability}
Understanding the practical relevance and comparability of different calibration metrics is essential when evaluating the QQU in regression models. Tab.~\ref{tab:metrics} provides an overview of the metrics introduced in the previous section, along with a summary of their key characteristics. In general, it is advantageous if a model expresses its uncertainty in the form of the parameters of a known PDF -- typically assuming a Gaussian distribution. From this, PIs can be easily derived by selecting an appropriate confidence level. The reverse is only possible when both the confidence level and the distributional form of the PI are known.  For example, considering a Gaussian distribution, a 95\% prediction interval approximately corresponds to the 2-$\sigma$ interval. Resulting in $\hat{y}_{\text{l},i} = \mu_i - 2\sigma_i$ and $\hat{y}_{\text{u},i} = \mu_i + 2\sigma_i$, with $\mu_i$ as the mean and $\sigma_i$ as the standard deviation of the predicted distribution.

A notable observation is that only the PICP, ECPE (which is derived from PICP) and CalS exclusively measure QQU. All other metrics capture a combination of QQU and predictive accuracy. This reflects the fact that most of these metrics originate from model recalibration settings, where the main goal is not only in improving the QQU, but preserving/improving prediction quality is vital as well. Thus, a model is considered well-calibrated if it achieves both high accuracy and meaningful uncertainty quantification. Consequently, calibration metrics are typically designed to reflect both properties simultaneously.

In practice, recalibration pipelines compare model performance before and after the application of a recalibration method. If the value of the metric improves, the recalibration is deemed successful. In this context, the absolute value of the metric and its interpretability are of secondary importance; what matters is the change in the metric. However, poor interpretability limits the usefulness of such metrics: typically, the only fixed point of reference is the perfectly calibrated model, often corresponding to a metric value of 0. Any other value provides little insight into how well-calibrated the model truly is, making it difficult to set specific target values as optimization goals -- unlike accuracy metrics such as the mean absolute error, where a given threshold can be meaningfully interpreted with respect to the target value domain. 

When comparing different recalibration approaches, the preferred method is usually the one that produces the largest improvement in the chosen metric. This practice assumes that all metrics respond consistently to improvements in model calibration. However, a closer look at the metrics reveals substantial differences in their formulation. For example, the CWC depends on a user-defined hyperparameter, the CRPS has the same unit as the target variable, and the NLL is influenced by the logarithm of the predicted variance. Thus, beyond differences in scale, these metrics also vary in how strongly they weight accuracy versus QQU.

Given this diversity, a critical question arises: do the metrics behave consistently across different recalibration scenarios? In other words, will a well-calibrated model always be rated higher by all metrics than its uncalibrated counterpart? If not, then it may be possible to present a model as well-calibrated simply by selecting a favorable metric -- without actually improving the underlying calibration. This raises the concern that calibration quality could be overstated based on metric choice alone. To address this issue, the following section introduces a benchmarking framework designed to investigate the consistency of these metrics under controlled conditions.

\begin{table}[htbp]
    \centering
    \caption{Summary table containing all presented metrics and their key properties.}
    \begin{tabularx}{\textwidth}{l c  c  p{1.6cm}  l  l  X}
    Metric & Accuracy & QQU & Uncertainty Type & Goal & Range & Comment\\
    \hline
    PICP &&\checkmark&Interval&Nominal&[0, 1]& Should match the nominal confidence level (e.g. 0.95)\\
    CWC &\checkmark&\checkmark&Interval& Decrease&[0, $\infty$) & Penalizes low PICP; Scale dependent on the hyperparameter $\eta$\\
    IS & \checkmark & \checkmark & Interval & Decrease & [0, $\infty$) & Proper scoring rule; Values have the same unit as the target variable\\
    CRPS &\checkmark&\checkmark&Density&Decrease&[0, $\infty$) &Proper scoring rule; Values have the same unit as the target variable\\
    NLL &\checkmark&\checkmark&Density&Decrease&(-$\infty$, $\infty$) &Proper scoring rule; depends on scale of target and log of predicted variance\\ 
    CalS &&\checkmark&Density&Decrease&[0, $\infty$) & Scale dependent on number of confidence thresholds $m$\\
    ENCE &\checkmark&\checkmark&Density&Decrease&[0, $\infty$) & Includes RMSE per bin, reflecting conditional accuracy\\
    ECPE &&\checkmark&Interval&Decrease&[0, 1] & Measures average deviation between PICP and nominal levels across intervals\\
    UCE &\checkmark&\checkmark&Density&Decrease&[0, $\infty$) & Weighted and scale-dependent on squared errors and predicted variances per bin.\\
    QCE & \checkmark &\checkmark &Density &Decrease & [0, 1] & Measures agreement between predicted uncertainty and empirical error frequencies\\
    \bottomrule
    \end{tabularx}
    \label{tab:metrics}
\end{table}

\section{Benchmark} \label{sec:benchmark}
To evaluate the consistency of different metrics for assessing model calibration, we conduct a comprehensive benchmark. A diverse selection of real-world and synthetic datasets is used to train data-driven models for predicting the respective target variable. For each of the models the performance is then assessed using a variety of calibration metrics. The results reveal substantial inconsistencies between metrics: for the same model, one metric may indicate high-quality uncertainty estimates, while another deems the same model's uncertainty poor.

The benchmark is structured in four parts:
\begin{itemize}
\item Evaluation of metric consistency on real-world datasets.
\item Evaluation of metric consistency on synthetic datasets.
\item Evaluation of metric consistency to recalibration.
\item Evaluation of metric sensitivity to controlled miscalibration.
\end{itemize}
To conclude the benchmark, a summary and application recommendations are provided at the end.

To evaluate the consistency of the metrics in the first two parts of the benchmark, we introduce an evaluation scheme based on the comparison of the normalized model performance. As outlined previously, the metrics differ substantially in scale and interpretation -- some being bounded, others semi-bounded (e.g., from 0 to $\infty$) or unbounded. This makes direct comparison of raw metric values across datasets unreliable. Furthermore, many metrics originate from recalibration literature, where relative changes (e.g., before vs. after recalibration) are the primary focus. To address these issues, we compute the metric values and normalize them by division with the mean value of said metric across all datasets. This preserves relative performance differences while bringing all metrics onto a comparable scale. For the PCIP we consider the absolute difference to the nominal confidence level. We then sort the datasets by performance and visualize the resulting values as line plots. This is inspired by the fact that a good model should be deemed good by all metrics. Ideally, consistent metrics should exhibit similar trends -- that is, their lines should follow the same general order across datasets. Strong deviations in direction or ordering (for instance, if a model performs well under one metric but poorly under another) indicate inconsistency and cast doubt on the reliability of the metric for evaluation of a model. In addition, we rank the datasets for each metric and assess the Spearman's rank correlation coefficient as consistent metrics should yield a high rank correlation. Both methods attest severe inconsistencies between metrics.

Motivated by these findings, the third part of the benchmark follows a more classical setup: we compare each model before and after recalibration using all metrics. The goal is not to determine whether recalibration improves model performance in general, but rather to illustrate how metric inconsistency affects the interpretation of recalibration results. The results demonstrate that the perceived effectiveness of recalibration depends heavily on the chosen metric -- highlighting how easily recalibration outcomes can be framed positively or negatively by selectively reporting favorable metrics.

To make a final statement about the usability of the metrics we have to compare the performance on data with known ground-truth uncertainty. Thus, in the third part, we evaluate how different calibration metrics respond to controlled perturbations in otherwise perfectly calibrated predictions. We first generate perfectly calibrated data by sampling predictions from a known Gaussian distribution centered at the true target value, with a known input-dependent standard deviation. Then, we apply targeted alterations to the predictions to introduce miscalibration, recompute the metrics, and compare the results to assess the metrics’ sensitivity and consistency under known conditions.

Overall, results confirm that substantial inconsistencies exist between the metrics. In the case of synthetic datasets and artificial miscalibration these inconsistencies persist, suggesting that the disagreement stems from the metrics themselves rather than from data-related noise or anomalies. However, we show that the CWC and ENCE come out superior in our benchmark.

The benchmark includes all metrics introduced in Section~\ref{sec:methods}. In addition, we evaluate RMSE, sharpness of the UQ and the Pinball loss, which corresponds to the Check score as defined by \citet{gneiting_strictly_2007}. Details on hyperparameters and software packages used are provided in Appendix~\ref{sec:appendix_implementation_details}.

As the predictive models, we use Deep Ensembles as proposed by \citet{Lakshminarayanan_simple}. Each ensemble consists of five submodels, each trained on a bootstrapped version of the training data. Every submodel is a multilayer perceptron with two hidden layers of 64 neurons each. We use a 70/30 train/test split and a batch size of 32. The model is trained to minimize the Negative Log-Likelihood (NLL) loss (see Section~\ref{sec:nll}), which allows each ensemble to output both a mean prediction and an associated predictive variance, assuming Gaussian-distributed uncertainty.

In total, we use 16 real-world and 10 synthetic datasets. References and details to the datasets are provided in Appendix~\ref{sec:appendix_implementation_details}.

All benchmark code is publicly available as open-source software on GitHub.\footnote{\url{https://github.com/JelkeWi/evaluating_the_quality_of_the_quantified_uncertainty}}

\subsection{Case study: real-world data} \label{sec:real_world}
In the first part of the benchmark, we assess the consistency of calibration metrics on 16 real-world datasets. These datasets vary in domain, size, and complexity, and reflect typical scenarios encountered in practical applications. For each dataset, we train a deep ensemble model and evaluate its performance using all metrics described previously. The aim is to examine whether the metrics agree in their assessment under realistic conditions.

\begin{figure}[htbp]
    \centering
    \includegraphics[width=0.9\textwidth]{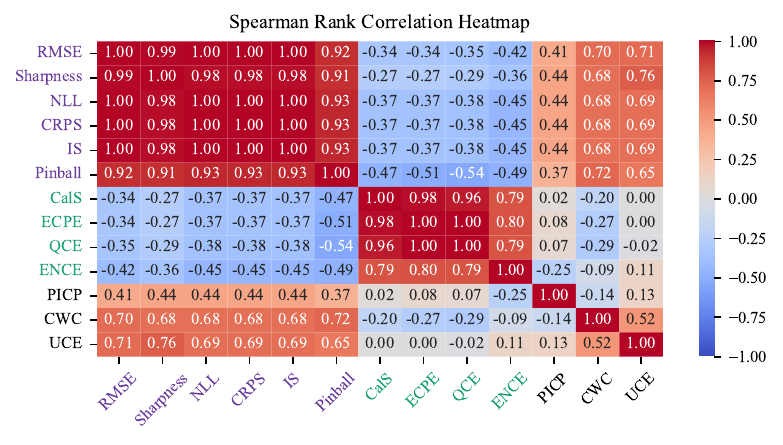}
    \caption{Heatmap of the Spearman's rank correlation coefficients using the real-world data. The evaluation metrics label color indicates the identified metric clusters: Purple for proper scoring rules and related metrics (Group 1), teal for threshold-based metrics (Group 2), and black for others (Group 3).}
    \label{fig:results_real_corr}
\end{figure}

The Spearman's rank correlation coefficients between the metrics' ranking of the datasets are presented as a heatmap in Fig.~\ref{fig:results_real_corr}. Comparing the correlation reveals two distinct groups that show internal agreement in how they assess model performance across the datasets.

The first and largest group -- located in the upper left corner -- consists of RMSE, Sharpness, NLL, CRPS, IS, and Pinball Loss. These metrics produce similar orderings of model performance and follow consistent trends across datasets. A commonality among them is that they are proper scoring rules and do not rely on discretizing the prediction or target space into bins. Instead, they directly assess the predictive distribution as a whole, including accuracy and uncertainty, leading to smoother and more globally consistent behavior. This theoretical foundation likely contributes to their empirical agreement.

A second distinct cluster comprises CalS, ECPE, QCE and ENCE. These metrics show strong agreement in their ordering of model performance, which can likely be attributed to their shared structural basis: all four rely on threshold-based and binned evaluation. CalS and ECPE assess model quality across multiple confidence levels, measuring whether predicted intervals correctly cover the target variable. QCE, while based on quantiles rather than confidence intervals, follows a similar logic by comparing predicted versus empirical quantiles across a spectrum of thresholds. The ENCE discretizes the target domain based on the standard deviations of the predictions of the model. This localized, thresholded evaluation approach distinguishes them from the first group of metrics and likely contributes to their internal consistency. However, it can be seen that the ENCE does not fully agree with the other metrics and the correlation is weaker.

The remaining metrics -- PICP, CWC, and UCE -- exhibit less consistent behavior. While there are partial overlaps or isolated agreements with the first group, they do not form a cohesive cluster. 

Importantly, even within the internally consistent groups, disagreements between groups are substantial. The contrasting ranking pattern between groups -- ranging from strong positive correlations to moderate negative correlations -- indicates that the order of datasets (from good to bad performance) is partially reversed.

\begin{figure}[htbp]
    \centering
    \includegraphics[width=1\textwidth]{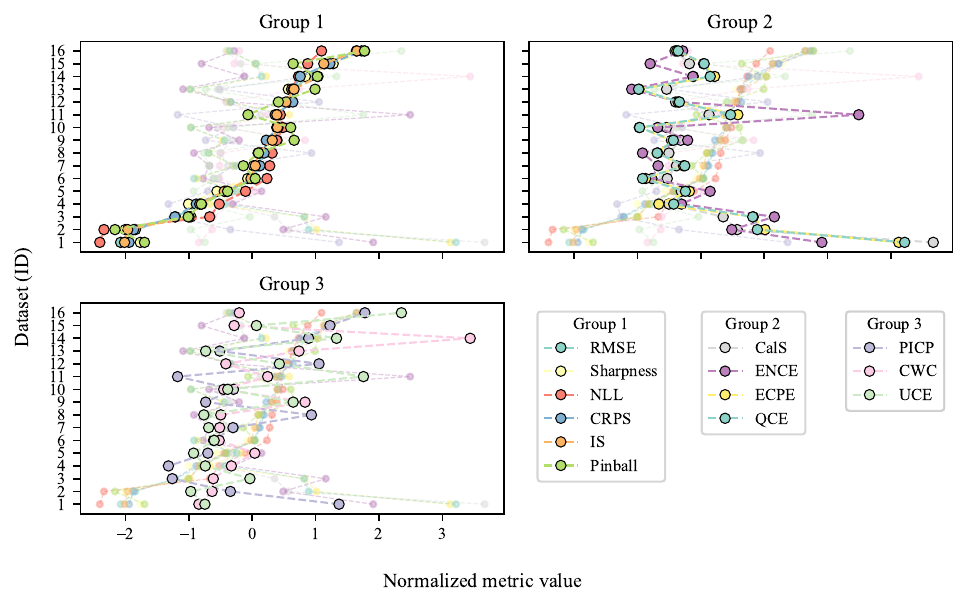}
    \caption{Normalized metric values for the deep ensembles across 16 real-world datasets. The values are normalized per metric by their mean across datasets. Groups 1 and 2 are formed based on similarity due to high correlation. Group 3 shows less/no internal consistency. Notably, metrics from different groups disagree on which model performs best on a given dataset, underscoring the substantial inconsistency in uncertainty evaluation. The IDs to the respective datasets are provided in Appendix \ref{sec:appendix_implementation_details}.}
    \label{fig:results_real}
\end{figure}

To gain more insights, we visually display the normalized metric values across the 16 real-world datasets in the identified groups (see Fig.~\ref{fig:results_real}). Overall, the visual impression supports the correlation and the contrasting ranking pattern between groups can be identified. For instance, in the case of the first dataset, the proper scoring rules (Group 1) consistently rate the ensemble model as highly reliable, while the threshold-based calibration metrics (Group 2) assign it the lowest performance. Such a reversal in evaluation highlights how the choice of metric can drastically influence conclusions, even when applied to the same model and data.

In the second case study, we investigate whether these inconsistencies persist under idealized conditions using synthetic datasets specifically designed to eliminate confounding factors such as aleatoric noise or irregular sampling.

\subsection{Case study: synthetic data} \label{sec:synthetic}
To isolate potential confounding factors such as aleatoric noise, outliers, or complex feature-target relationships, we repeat the analysis using synthetic datasets. These datasets are generated with fully controlled data-generating processes and are designed to be well-behaved in the sense that they exhibit smooth, noise-free functional relationships between input features and the target variable. Specifically, we exclude both stochastic noise in the target and structural irregularities in the feature space that could bias uncertainty estimation. This setup allows us to test the metrics under idealized conditions, where the ground-truth uncertainty is fully determined by the model’s epistemic uncertainty (inductive bias and data coverage), rather than aleatoric uncertainty (external sources of variability).

By evaluating metric agreement in this controlled setting, we can rule out unpredictable behavior in the data as a cause for the inconsistency observed in the real-world case. If inconsistencies persist in the absence of aleatoric uncertainty or data artifacts, this strongly suggests that the disagreement stems from inherent properties and sensitivities of the metrics.

\begin{figure}[htbp]
    \centering
    \includegraphics[width=0.9\textwidth]{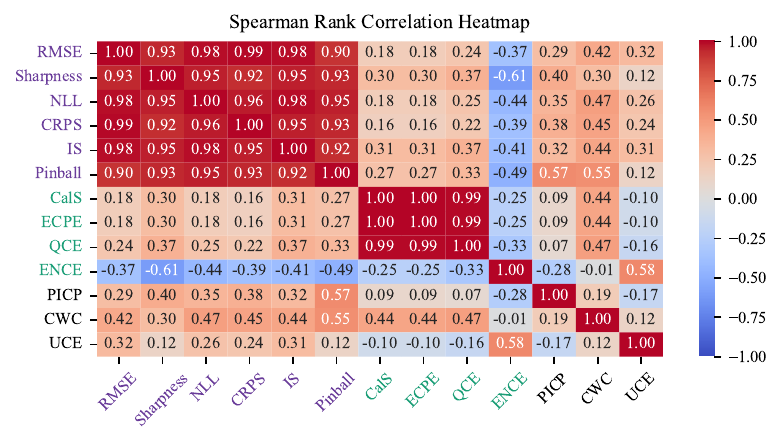}
    \caption{Heatmap of the Spearman's rank correlation coefficients using synthetic data. The evaluation metrics label color indicates the previously identified metric cluster: Purple for proper scoring rules and related metrics (Group 1), teal for threshold-based metrics (Group 2), and black for others (Group 3).}
    \label{fig:result_synthetic_corr}
\end{figure}

The Spearman's rank correlation heatmap on the synthetic datasets (see Fig.~\ref{fig:result_synthetic_corr}) show a similar clustering of metrics to that observed for the real-world data, albeit the contrasting ranking between groups has partly diminished. Most metrics now exhibit strong to no correlation, but no negative values -- apart from the ENCE. Nevertheless, the groups still exist. The ENCE, however, shows weak to moderate negative correlations with most other metrics, thereby ranking model performance in the opposite order.

\begin{figure}[htbp]
    \centering
    \includegraphics[width=\textwidth]{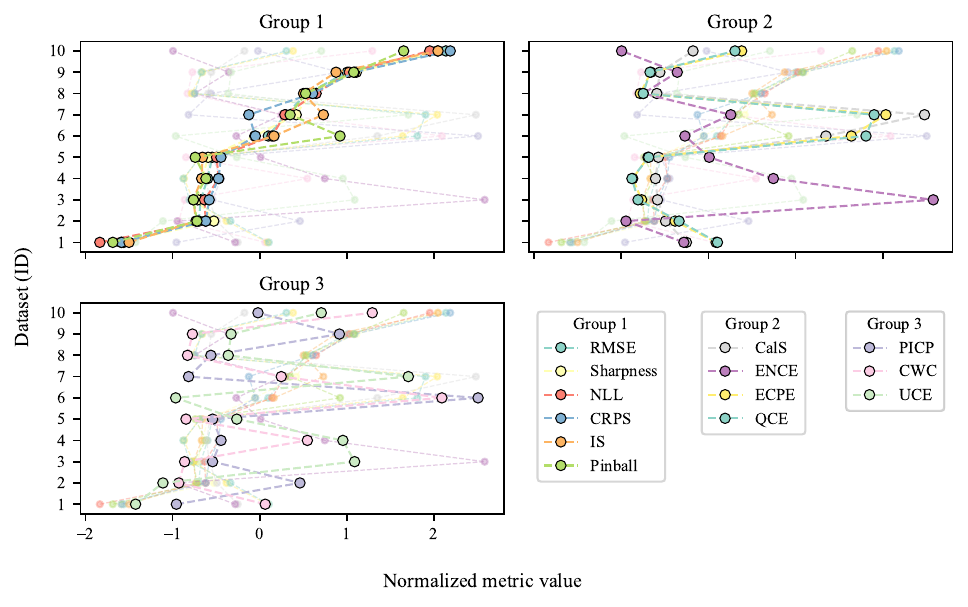}
    \caption{Normalized metric values for the deep ensembles across 10 synthetic datasets. The values are normalized per metric by their mean across datasets. The persistence of the three groups, despite the absence of aleatoric noise in the synthetic datasets, suggests that the inconsistencies are due to intrinsic differences in metric definitions. This emphasizes the need for caution when selecting metrics to evaluate uncertainty quality. The IDs to the respective datasets are provided in Appendix \ref{sec:appendix_implementation_details}.}
    \label{fig:result_synthetic}
\end{figure}

The visual inspection of the metric values (see Fig.~\ref{fig:result_synthetic}) confirms the clustering of metrics -- again with the exception of the ENCE.

Overall, the benchmark indicates that the inconsistencies between metric groups are not caused by data noise or aleatoric uncertainty, since the synthetic datasets are specifically constructed to be noise-free and well-behaved. Instead, the observed disagreements appear to stem from fundamental differences in what the metrics evaluate, reinforcing the findings from the real-world benchmark.

The third part of the benchmark further demonstrates that these inconsistencies have a major impact on the evaluation of recalibration methods, as the perceived effectiveness of recalibration varies strongly depending on the chosen metric.

\subsection{Case study: recalibration}\label{sec:calibration}
Building on the findings from the previous parts, this case study investigates how metric inconsistency affects the evaluation of model recalibration. For each real-world dataset, we compare the performance of models before and after applying a recalibration method, using all previously introduced metrics. The goal is not to determine whether recalibration improves model quality per se, but rather to assess how the perceived effect of recalibration depends on the metric used. This highlights the practical implications of metric inconsistency: depending on the choice of metric, the same recalibration procedure can appear beneficial, neutral, or even detrimental.

To conduct this analysis, we apply four established recalibration methods:
\begin{itemize}
    \item Isotonic Regression \cite{kuleshov_accurate_2018}, a non-parametric technique that recalibrates the cumulative distribution function of predictions.
    \item Variance Scaling \cite{levi_evaluating_2022, laves_well-calibrated_2020}, which rescales predicted variances to better align predicted and empirical uncertainty.
    \item GP Normal \cite{kuppers_parametric_2023}, a parametric method that fits Gaussian distributions to the predicted uncertainty.
    \item GP Beta \cite{song_distribution_2019}, which recalibrates uncertainty estimates using Beta-distributed predictive intervals.
\end{itemize}

Rather than focusing on the absolute metric values, which may be difficult to interpret due to differing scales and directions, we compute the relative change in each metric (in percent) after recalibration. This allows for a standardized comparison across metrics. For all metrics (except the PICP) lower values indicate better performance. Thus, a negative relative change implies an improvement in calibration due to recalibration. For the PICP, which measures how closely the empirical coverage matches a nominal confidence level, we instead compute the absolute deviation from the nominal level. Here, a negative change also indicates improvement, i.e., the coverage moving closer to the desired target. 

To avoid over-interpreting fluctuations caused by numerical noise or negligible recalibration effects, we report only those changes where the relative difference exceeds 3\%.

By analyzing the direction and magnitude of metric changes under each recalibration method, we further illustrate how inconsistent metrics can lead to contradictory conclusions. This reinforces the need for caution when evaluating recalibration techniques based on a single metric, as such evaluations may be arbitrarily influenced by the choice of metric rather than the true improvement in model reliability.

\begin{figure}[htbp]
    \centering
    \includegraphics[width=0.9\textwidth]{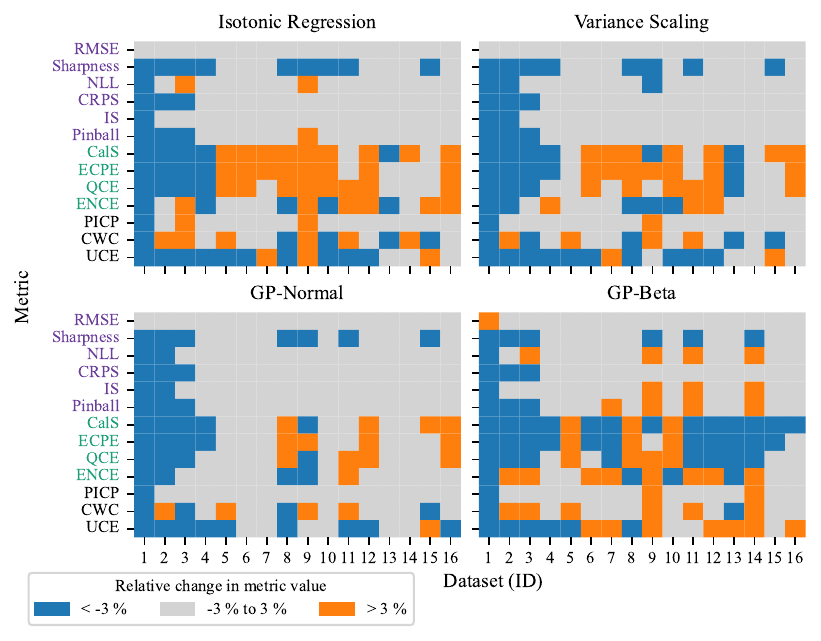}
    \caption{Relative change in metric value after model recalibration. Each heatmap shows whether a metric detects a substantial improvement (blue), degradation (orange), or negligible change (grey) in model performance after calibration. The evaluation metrics label color indicates the previously identified metric cluster: Purple for proper scoring rules and related metrics (Group 1), teal for threshold-based metrics (Group 2), and black for others (Group 3).}
    \label{fig:results_calibration}
\end{figure}

The results are illustrated in Fig.~\ref{fig:results_calibration}. Ideally, consistent metrics would yield uniformly colored columns in the matrix, indicating agreement on whether recalibration improved or worsened performance. Ideally, we would also uniformly assume improvement of calibration after recalibration. However, this is not the case. Instead, we observe a clear split between the two previously identified metric groups: within each group, the metrics tend to agree on the direction of change, but between groups, their assessments often diverge substantially. Notably, the proper scoring rules (Group 1) frequently show little or no change after recalibration, indicating that they are less sensitive to adjustments of the quantified uncertainty introduced by the recalibration methods. In contrast, metrics in Group 2 (based on threshold-based or confidence-level comparisons) tend to report more pronounced changes, both positive and negative. 

Interestingly, while recalibration often improved uncertainty sharpness, these gains were not always reflected in proper scoring rules like CRPS or NLL, which are commonly used in practice.

Crucially, for nearly every dataset and recalibration method, at least one metric reports an improvement while another indicates a deterioration. This inconsistency highlights a serious challenge: the effectiveness of a recalibration method can appear entirely different depending on which calibration metric is used. Furthermore, a non-negligible number of recalibrations even lead to a decrease in model performance across several metrics, suggesting that recalibration is not universally beneficial in practice -- although evaluating this properly is complicated by the very inconsistencies demonstrated here and the lack of ground truth uncertainties. In the following, we therefore conduct a test with controlled miscalibration, where the ground truth is known.

\subsection{Case study: controlled miscalibration} \label{sec:miscalibration}
In the previous three parts of the benchmark, the lack of ground-truth uncertainty posed a fundamental limitation: without knowing the true uncertainty, it is not possible to determine which calibration metric most accurately reflects the model’s reliability. To address this, the fourth part of our benchmark evaluates how calibration metrics respond to controlled, synthetic miscalibrations in otherwise perfectly calibrated predictions.

We generate near-realistic data by using the real-world regression datasets as a basis, but simulate perfectly calibrated predictions. Specifically, we assume predictions are drawn from a Gaussian distribution centered at the true target value $y$, with an input-dependent standard deviation $\hat{\sigma}$ composed of a structured (epistemic) and unstructured (aleatoric) component:
\begin{align}
    \sigma_{\text{epistemic}} &= 0.05 \cdot R + \sin^2\left( \frac{2\pi y}{R}\right),\\
    \sigma_{\text{aleatoric}} &= \mathcal{N}\left(0, \left(0.05\cdot R\right)^2\right),\\
    \hat{\sigma} &= \text{max}\left(\sigma_{\text{epistemic}} + \sigma_\text{aleatoric}, \epsilon\right),\\
    \hat{y} &\sim \mathcal{N}(y, \hat{\sigma}^2).
\end{align}
where $R$ is the range of target values and $\epsilon$ is a small constant to avoid degenerate values. This construction yields sinusoidal heteroscedastic predictive uncertainty with both structured and random components, approximating realistic modeling conditions while guaranteeing perfect calibration.

We first evaluate all metrics on these perfectly calibrated predictions to establish a baseline. We then introduce artificial miscalibration through four distinct perturbation scenarios and re-evaluate the metrics. Each scenario targets a different aspect of the predictive distribution:

\begin{itemize}
    \item Constant standard deviation offset: All predicted standard deviations $\hat{\sigma}$ are scaled by a fixed factor of 0.9, simulating systematic underestimation of uncertainty.
    \item Heterogeneous standard deviation offset: Standard deviations are scaled by factors linearly varying between 0.9 and 1.1 across samples, introducing input-dependent uncertainty distortion.
    \item Constant mean prediction offset: Predicted means $\hat{y}$ are scaled by a factor of 0.9, resulting in biased point predictions.
    \item Heterogeneous offset of both mean and standard deviation: Mean predictions are scaled by factors from 0.9 to 1.1, while standard deviations are inversely scaled from 1.1 to 0.9, reflecting realistic and adversarial degradation in both components.
\end{itemize}

\begin{figure}[htbp]
\centering
\includegraphics[width=0.9\textwidth]{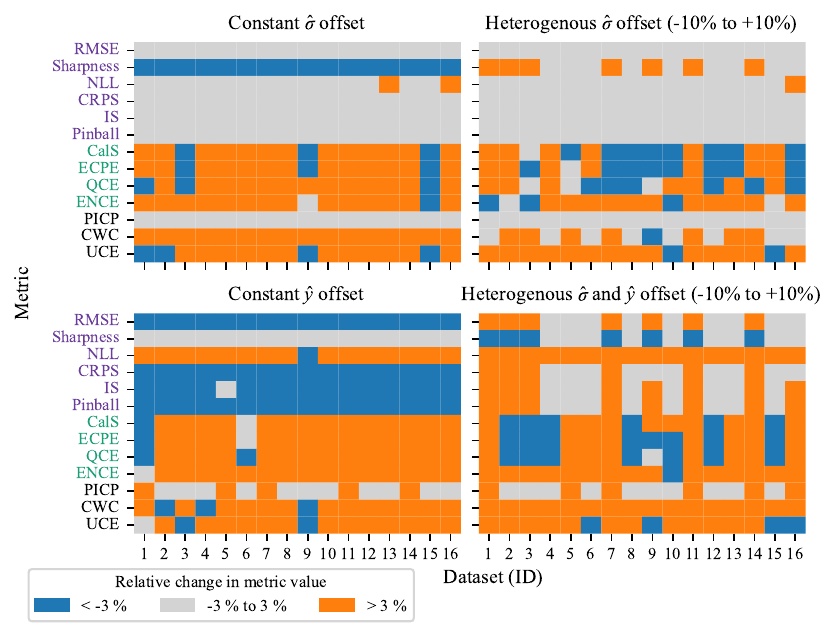}
\caption{Relative change in metric values after artificially miscalibrating predictions across four scenarios. Each heatmap cell indicates whether a metric detects an improvement (blue), degradation (orange), or negligible change (grey). It would be correct if a degradation is identified. Metrics are colored by their previously identified group: purple (Group 1), teal (Group 2), black (Group 3).}
\label{fig:result_miscalibration_single}
\end{figure}

As in the recalibration case study (Section~\ref{sec:calibration}), we focus on changes in metric values exceeding a relative threshold of 3\%. Fig.~\ref{fig:result_miscalibration_single} shows how each metric responds to the different miscalibration types. Metrics in Group 2 (threshold-based metrics) almost consistently detect the constant $\hat{\sigma}$ offset, while metrics in Group 1 (proper scoring rules) generally do not. The heterogeneous $\hat{\sigma}$ offset yields less consistent results, with some Group 2 metrics misidentifying the degraded data as better calibrated, highlighting their sensitivity but also potential unreliability. Here we can see once again the outstanding performance of ENCE in group 2, which usually correctly detects degradation. The constant $\hat{y}$ offset scenario is particularly revealing: among the proper scoring rules, only the NLL reliably flags the miscalibration, while others misinterpret the degraded predictions as improved. In contrast, most metrics from Groups 2 and 3 correctly identify the calibration deterioration.

The fourth and most complex scenario with simultaneous heterogeneous offset of both mean and standard deviation exposes further nuances. While NLL again stands out within Group 1 for its consistent detection, other proper scoring rules show limited sensitivity. Group 2 metrics detect changes in calibration across the board, but for about half of the datasets erroneously an improvement is detected, suggesting poor alignment with ground-truth reliability. Notably, the ENCE quite consistently detects the miscalibration together with metrics such as CWC and UCE -- which previously did not cluster clearly with other groups -- and thus emerge as the most consistent, correctly identifying degradation in nearly all datasets and scenarios.

Because some real-world datasets are small (fewer than 200 samples), the random sampling of predictions and binning used in some metrics can introduce high variability. This results in instability across benchmark runs. To account for this, we repeat the entire miscalibration benchmark 100 times and count how often each metric correctly detects degradation (defined as a $\geq3\%$ worse calibration score) in the fourth, most realistic, scenario.

\begin{figure}
\centering
\includegraphics[width=0.9\textwidth]{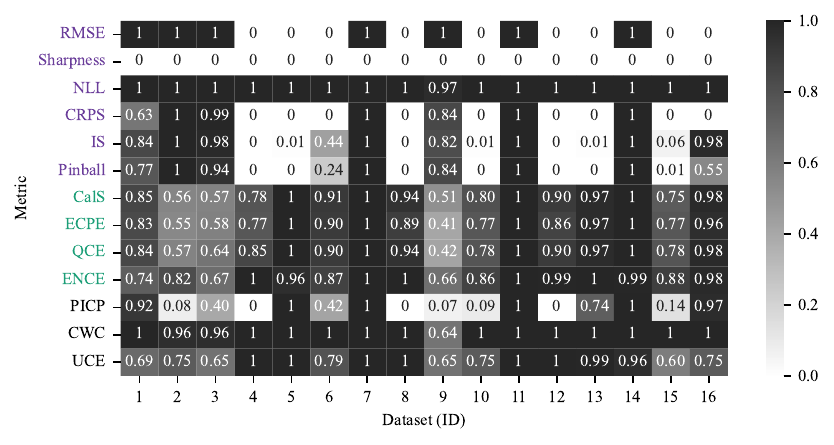}
\caption{Stochastic evaluation of metric sensitivity under controlled miscalibration. The heatmap shows the fraction of runs (out of 100) in which each metric correctly detected the artificial miscalibration in the fourth scenario -- heterogeneous offsets applied to both the predicted mean and standard deviation. A miscalibration is deemed correctly identified if the relative change in metric value exceeds 3\%. The metric names are colored by their previously identified group: purple (Group 1), teal (Group 2), black (Group 3).}
\label{fig:result_miscalibration_stochastic}
\end{figure}

Fig.~\ref{fig:result_miscalibration_stochastic} summarizes these results. The ability to detect miscalibration varies substantially between datasets. For example, datasets 3 and 9, which have the fewest samples, exhibit the lowest detection frequencies -- highlighting the impact of sample size on calibration evaluation reliability. As dataset size increases, so does the likelihood of correct detection. Among the proper scoring rules, only NLL reliably flags the miscalibration. Group 2 metrics behave similarly (group internal) but remain sensitive to noise in smaller datasets.

Overall, the CWC, ENCE, and UCE again stand out for their robustness. Despite lacking clear internal consistency across previous benchmark parts, they consistently detect miscalibration in this setting and outperform Group 2. The ENCE is marginally more reliable than UCE, while CWC appears to be the most stable across datasets and runs.

\subsection{Recommendations}\label{sec:recommendations}
Our results suggest that calibration metrics provide consistent results when evaluated on datasets with at least 500–1000 samples, assuming a typical number of bins for binning-based approaches (e.g., 10 bins). For smaller datasets, stochastic effects (such as sampling variability and binning noise) can lead to unstable or contradictory results, regardless of whether these conclusions reflect the true calibration quality.

Among all evaluated metrics, the NLL, CWC and ENCE stand out as particularly informative.

The NLL consistently detects miscalibration when both the predicted mean and the uncertainty are perturbed. However, it is also shown to be largely insensitive to changes in the predicted uncertainty alone, making it unsuitable as a comprehensive calibration metric that reflects both accuracy and uncertainty quality.

The CWC emerged as the most reliable metric across all benchmarks. It extends the classical prediction interval coverage probability (PICP) by incorporating the width of the prediction interval, and is evaluated at a fixed nominal confidence level. Despite its strong performance, it comes with disadvantages. Firstly, it relies on a user-defined penalty parameter $\eta$ (set to 50 in our study) which introduces an additional hyperparameter that can influence results. Secondly, the CWC focuses only on a single confidence level, thus in domains with heteroscedastic uncertainty it may overlook local miscalibrations if badly calibrated regions are balanced out by well-calibrated regions. And thirdly, assessing only one confidence level limits the expressiveness to said confidence level (e.g. 95\%), which not necessarily also implies that the model is calibrated for other levels as well. To mitigate this problem, one could consider calculating the mean CWC across multiple confidence levels. However, the exponential term would potentially skew the mean.

The ENCE performed well, demonstrating consistent sensitivity to miscalibration in both mean and uncertainty, while not requiring any hyperparameters. Unlike the CWC, ENCE evaluates calibration across multiple bins and is thus better suited to detect local patterns of miscalibration. It also avoids the ambiguity of choosing a specific confidence level and can provide a more fine-grained view of calibration quality. However, for small datasets, the ENCE does not always correctly detect miscalibration. Nevertheless, it still outperforms the other metrics of the benchmark.

We therefore recommend ENCE as a practical and robust choice for evaluating predictive uncertainty calibration, particularly due to its lack of required hyperparameters, consistent performance across datasets and scenarios, and ability to capture miscalibration both globally and locally.

\section{Discussion} \label{sec:discussion}
In the previous sections, calibration metrics were presented and evaluated. However, the approach chosen here is by no means beyond doubt, which is why we would like to put some points up for discussion.

We have introduced a taxonomy that describes calibration metrics as metrics that capture the QQU and accuracy of a model. In theory, for most recalibration procedures, capturing the QQU alone would suffice, but potential influences on accuracy would then not be captured. We therefore consider the combination of the two performance indicators in a calibration metric to be appropriate. This is particularly relevant in view of potential future recalibration procedures that influence both.

Most metrics in our benchmark implicitly assume that predictive uncertainties follow a Gaussian distribution. While this assumption is common and facilitates tractable evaluation, it may not hold in practice. Nonetheless, it reflects the minimal modeling assumption required to interpret uncertainty quantities consistently across methods.

Although the experiments were conducted using deep neural networks in an ensemble configuration, the calibration metrics and recalibration methods themselves are model-agnostic. They do not rely on specific properties of neural networks. We therefore expect the results to generalize beyond the specific model type used and any specific method to obtain uncertainty estimates.

Another important consideration is that we did not assess the initial calibration state of models prior to recalibration and thus might have tried to recalibrate an already calibrated model. As the metric values are often non-interpretable on their own, this is a common circumstance. For the purpose of comparing metric behaviors, this is inconsequential: regardless of whether recalibration improves the predictive quality of the model, the inconsistency in the responses of the metrics remain. 

Further, while we used 10 bins for metrics based on thresholding or binning (e.g., ECPE, QCE), we acknowledge that the bin count may influence results. We repeated the analysis with 7 and 15 bins and observed no substantial changes in metric behavior, suggesting relative robustness in this setting. However, the amount of bins is dictated by the number of samples as bin-dependent metrics inherently introduce discretization artifacts and hyperparameter sensitivity that should be considered when interpreting results. We found that on average 100 samples per bin suffice for the evaluated metrics. Albeit, this is just an educated guess and is also dependent on the variability of the data.

Calibration metrics typically assess the average alignment between predicted uncertainty and observed errors across the entire dataset. However, this aggregate view can obscure poor calibration at the level of individual samples. As noted by \cite{zhao_individual_2020, sluijterman_howtoevaluate}, a model may appear well-calibrated on average while systematically under- or overestimating uncertainty for certain subgroups or regions of the input space. Using more bins for discretization breaks down the issue into smaller sections, but it still persists. This closely parallels challenges in imbalanced regression, where performance may be high overall but degraded in underrepresented target ranges, cf.~\cite{wibbeke_model-agnostic_2025}. Such cases highlight the importance of analyzing calibration beyond global summaries, especially in applications where reliable predictions are needed for rare or critical instances.

Finally, the set of metrics considered in this study is not exhaustive. For instance, \citet{kristoffersson_lind_uncertainty_2024} explored metrics such as Spearman rank correlation between the predicted uncertainty and the error, but deemed it inferior to CalS, which is why we refrained from including it.

\section{Outlook} \label{sec:outlook}
This benchmark study highlights critical inconsistencies and limitations in how calibration is evaluated in regression models but intentionally refrains from proposing definitive solutions. Our goal is to surface these challenges and spark a broader discussion on the need for more reliable evaluation frameworks.

A major limitation remains the lack of ground truth for uncertainty. Since true predictive uncertainty is inherently unobservable, it is difficult to determine which calibration metric best reflects the actual reliability of model predictions. While existing metrics offer useful proxies, they cannot be directly validated. We tried to address this by introducing artificial miscalibration, however, simulation-based approaches as proposed by \cite{sluijterman_howtoevaluate} may provide a promising direction. Such setups allow full control over the data-generating process and enable direct comparison between predicted and true uncertainty. Potentially this could guide the development of more principled metrics, which are evaluated on synthetic procedures and later on applied to real-world problems.

Another promising avenue stems from the field of imbalanced regression, where relevance functions are used to assign importance to different samples based on their target values. A similar concept could be adopted in uncertainty evaluation to weight samples according to how well their individual uncertainty is captured, especially in cases of heteroscedastic or distributionally skewed data. This would help move beyond purely aggregate assessments and toward more nuanced, localized evaluations of calibration quality.

Similarly, \citet{wibbeke_model-agnostic_2025} showed that ensembles can help mitigating model bias caused by data imbalance. Thus, when already forming an ensemble to avoid bias, one could also think of using the standard deviation of the models as a measure of uncertainty.

Ultimately, future work should aim not only to refine existing metrics, but also to establish stronger theoretical and empirical foundations for calibration evaluation, ideally grounded in interpretable criteria, robustness across domains, and avoidance of binning.

\section{Conclusion} \label{sec:conclusion}
This work addressed a critical yet underexplored aspect of uncertainty estimation in regression: the evaluation of calibration. While predictive models are increasingly used in applications, the trustworthiness of their uncertainty estimates remains difficult to assess. To tackle this challenge, we conducted a comprehensive investigation into calibration metrics, which are essential for performance evaluation, e.g. improving the quality of uncertainty estimates through recalibration.

We began by clarifying the often-confused terminology surrounding calibration and recalibration, providing a clear taxonomy that distinguishes between the state of being calibrated (accurate predictions and reliable uncertainty) and the process of recalibration (post-hoc improvement of this state). From the literature, we extracted a wide range of calibration metrics and categorized them by conceptual foundations. Our analysis revealed that these metrics differ not only in their focus (some targeting accuracy, others uncertainty), but also in scale, assumptions, and interpretability, making direct comparisons difficult.

Through a rigorous benchmark using real-world, synthetic and synthetically miscalibrated data, we demonstrated that calibration metrics frequently produce conflicting evaluations. For any given recalibration method, it is often possible to find a metric that reports improvement and another that suggests degradation. This inconsistency is particularly concerning as it allows cherry-picking of metrics to create misleading impressions of success. Even within conceptually related groups, such as proper scoring rules, internal agreement does not guarantee correctness. Importantly, we found that the ENCE and CWC metrics showed superior behavior in detecting miscalibration, though ENCE benefits from structural advantages. Furthermore, our results revealed that calibration metrics become unreliable on small datasets, typically requiring at least 500–1000 samples to yield meaningful results.

Overall, this study highlights the fragility and ambiguity of current calibration evaluation practices and emphasizes the importance of metric choice when assessing recalibration effectiveness. Our findings suggest that relying on a single metric can produce biased conclusions and that multiple, theoretically grounded metrics should be reported to provide a more robust evaluation. By scrutinizing the evaluation metrics themselves, this work lays the foundation for more trustworthy and reproducible research in uncertainty quantification and model recalibration.

Beyond our empirical findings, this study highlights that calibration evaluation is fundamentally limited by the unobservable nature of true predictive uncertainty. Since this cannot be circumvented in real-world applications, simulation-based testing with synthetic data may provide the most promising avenue to evaluate and compare metric performance. Moreover, as metrics and recalibration methods are model-agnostic, we expect our results to generalize beyond neural networks, but future work should aim for more localized and theoretically grounded evaluation criteria.

\bibliographystyle{unsrtnat}
\bibliography{references}  %%% Uncomment this line and comment out the ``thebibliography'' section below to use the external .bib file (using bibtex) .

\appendix

\section{Implementation Details} \label{sec:appendix_implementation_details}
In this section information about the used datasets and hyperparameters as well as implementation details are provided.

For the PICP, ENCE, UCE, QCE and Pinball loss the python package ``net:cal - Uncertainty Calibration''\footnote{\url{https://github.com/EFS-OpenSource/calibration-framework}} of \citet{kuppers_parametric_2023} is used. For the CRPS, NLL and sharpness the Uncertainty Toolbox\footnote{\url{https://uncertainty-toolbox.github.io/}} by \citet{chung_uncertainty_2021} is used. All metrics (e.g., ENCE, ECE, QCE, and Pinball loss) that perform some kind of subdivision/discretization of the domain (e.g., dividing uncertainty into bins) divide the value space into 10 parts. The PICP assumes a confidence level of 0.95. The CWC assumes $\eta = 50$. For all other hyperparameters the default arguments of the respective software package are used.

A list of the real-world datasets used is provided in Tab.~\ref{tab:datasets}. All datasets are available in the UC Irvine Machine Learning Repository\footnote{\url{https://archive.ics.uci.edu/}}. Only samples without missing values are considered.

A list of all synthetic datasets used is provided in Tab.~\ref{tab:synthetic_data}. Some of the synthetic datasets are created using the sklearn library \citep{scikit-learn}. All synthetic datasets consist of 1000 samples and are generated without noise.

For both dataset types, all features are normalized by minmax-scaling between 0 and 1. To avoid model distortions due to exceeding noise or erroneous samples, all datasets are cleaned of outliers using an isolation forest outlier detection with a probability threshold of 0.8 \citep{liu2012isolation}. 

\begin{table*}[tb] % the * indicates a two-column stretch for 2-col layouts
\caption{List of used real-world datasets.}
\centering
\begin{tabular}{rl>{\raggedleft\arraybackslash}p{3cm}l}
No. (ID) & Name & {Number of samples after cleaning} & Ref. \\
\hline
1 & forest-fires & 513 &\citep{data_forest_fires_162} \\
2 & facebook-metrics &493& \citep{data_facebook_metrics_368} \\
3 & computer-hardware &206& \citep{data_computer_hardware_29} \\
4 & abalone & 4145&\citep{data_abalone} \\
5 & winequality-white &6494& \citep{Cortez2009ModelingWP} \\
6 & airfoil-self-noise &1488& \citep{data_airfoil_self-noise_291} \\
7 & superconductivity-data &20926& \citep{misc_superconductivty_data_464} \\
8 & grid-stability &9931& \citep{data_electrical_grid_stability_simulated_data__471} \\
9 & servo &164& \citep{data_servo_87} \\
10 & concrete-compressive-strength & 1010& \citep{data_concrete_compressive_strength_165} \\
11 & steel-industry & 34941& \citep{VE2020EfficientEC} \\
12 & combined-cycle-power-plant & 9547&\citep{data_combined_cycle_power_plant_294} \\
13 & parkinson & 5821&\citep{tsanas2009accurate} \\
14 & auction-verification & 1980& \citep{ordoni2022analyzing}\\
15 & energy-efficiency & 757 &\citep{misc_energy_efficiency_242}\\
16 & age-prediction &2267 &\citep{Dinh2019ADA} \\
\end{tabular}
    \label{tab:datasets}
\end{table*}

\begin{table}[tb]
\centering
\caption{List of generated synthetic datasets. *: \url{https://scikit-learn.org}}
\label{tab:synthetic_data}
\renewcommand{\arraystretch}{1.5}
\begin{tabularx}{\textwidth}{rlrX}
No. & Name & Features & Description \\
\hline
1 & Euclidean distance & 2 &  $y= \sqrt{a^2 + b^2}$\\
2 & Nernst equation & 6 &  $y= \frac{R \cdot T}{z \cdot F} \cdot \text{log}\left(\frac{a_1}{a_2}\right)$\\
3 & Stribeck translational friction & 6 &  $y= \mu_1 \cdot F + \left(\mu_2 \cdot F + \mu_1 \cdot F \right)\cdot \text{exp}\left(- |\frac{\vartheta_1}{\vartheta_2}|^\delta\right)$\\
4 & Inverse tangens & 1 &  $y= \text{arctan}(x)$\\
5 & Multilayer perceptron model & 4 & Random initialisation of a MLP model with gaussian weights.\\
6 & Random linear model & 3 &  Dataset using the \verb|make_regression| function of sklearn*.\\
7 & Sparse uncorrelated & 4 &  Dataset using the \verb|make_sparse_uncorrelated| function of sklearn*.\\
8 & Friedman1 & 5 &  Dataset using the \verb|make_friedman1| function of sklearn*.\\
9 & Friedman2 & 4 &  Dataset using the \verb|make_friedman2| function of sklearn*.\\
10 & Friedman3 & 4 &  Dataset using the \verb|make_friedman2| function of sklearn*.\\
\end{tabularx}
\end{table}

\end{document}